\documentclass[conference]{IEEEtran}
\IEEEoverridecommandlockouts
\usepackage{amsmath,amssymb,amsfonts}
\usepackage{algorithmic}
\usepackage{graphicx}
\usepackage{textcomp}
\usepackage{xcolor}
\usepackage{multirow}
\usepackage{booktabs}   
\usepackage{makecell} 
\usepackage{pgfplots}
\pgfplotsset{compat=1.18}

\usepackage[backend=biber,
            style=ieee,
            doi=true,
            url=false,
            isbn=false,
            sorting=none]{biblatex}

\usepackage[ruled,vlined]{algorithm2e}
\def\BibTeX{{\rm B\kern-.05em{\sc i\kern-.025em b}\kern-.08em
    T\kern-.1667em\lower.7ex\hbox{E}\kern-.125emX}}

\begin{document}

\title{Weather-Aware Domain Adaptation for Street-View Weather Recognition\\
}

\author{\IEEEauthorblockN{Hossein Maghsoumi$^1$, George Atia$^{1,2}$, Yaser P. Fallah$^1$}
\IEEEauthorblockA{$^{1}$\textit{Department of Electrical and Computer Engineering} \\
$^{2}$\textit{Department of Computer Science}\\
\textit{University of Central Florida,
Orlando, USA} \\
Emails: \{hossein.maghsoumi, george.atia, yaser.fallah\}@ucf.edu}
}

\maketitle

\begingroup
\renewcommand\thefootnote{}
\footnotetext{\footnotesize
\textcopyright~2026 IEEE. Personal use of this material is permitted.  Permission from IEEE must be obtained for all other uses, in any current or future media, including reprinting/republishing this material for advertising or promotional purposes, creating new collective works, for resale or redistribution to servers or lists, or reuse of any copyrighted component of this work in other works.

This is the accepted version of the paper published in the
2026 IEEE Conference on Technologies for Sustainability (SusTech).
DOI: 10.1109/SUSTECH67720.2026.11536286.
}
\addtocounter{footnote}{-1}
\endgroup

\begin{abstract}
Adverse conditions such as rain, snow, fog, and dust remain challenging for camera-based perception in autonomous driving. We study multi-class weather recognition from street-view images under domain shift, where most available training data come from non–street-view sources that differ markedly from real driving scenes. We propose Weather-Aware Adversarial Discriminative Domain Adaptation (WA-ADDA), which conditions the domain discriminator on predicted weather to promote features that are both domain-invariant and weather-sensitive. We also assemble a multi-dataset benchmark by unifying diverse non–street-view weather collections as sources and real street-view images as targets, and define a standardized evaluation protocol with macro accuracy as the primary metric. Across backbones (ResNet-50, EfficientNet, VGG, DenseNet), WA-ADDA consistently improves street-view performance and yields strong per-class recalls in challenging conditions while preserving clear-weather accuracy. These findings highlight the feasibility of domain-adapted weather recognition and the value of our benchmark for advancing robust, on-board perception.
\end{abstract}

\begin{IEEEkeywords}
Weather Classification, Autonomous Driving, Domain Adaptation, street-view recognition, WA-ADDA.
\end{IEEEkeywords}

\section{Introduction}
Autonomous driving spans levels from driver assistance to full autonomy, where safe operation hinges on reliable perception. Among on-board sensors, cameras are central: they capture rich scene cues used for detection, tracking, and control. Yet camera-based perception degrades under adverse weather. Rain, snow, fog, and dust reduce contrast and visibility, obscure road markings, and introduce artifacts (e.g., raindrops on windshields, specular glare), degrading object detection and lane following. Automatically recognizing weather from images is thus a practical perception subtask that can support downstream safety logic (e.g., adaptive speed policies, sensor-fusion reweighting, and dynamic sensitivity of pedestrian and lane detectors).

External weather services (global feeds, satellite/radar) are invaluable but often coarse in space and time and may miss \emph{on-scene} phenomena critical to vision: local fog banks, brief dust plumes, wet-but-not-raining pavement, or residual snow on the road despite no active snowfall. In contrast, the ego camera provides \emph{immediate, localized} evidence even when connectivity is limited. Therefore, image-based weather recognition is essential for on-board perception, enabling timely, localized adaptation of detection and control.

Despite substantial work on detection and segmentation in adverse weather, direct multi-class weather recognition for street-view images remains relatively underexplored. Most available datasets are non–street-view or synthetically augmented, and domain shift—differences in viewpoint, optics, and scene composition—causes models trained on one dataset to generalize poorly to real driving imagery.

Domain adaptation (DA) addresses such shifts by aligning feature distributions between labeled sources and unlabeled targets. Adversarial Discriminative Domain Adaptation (ADDA) is a widely used two-stage approach: supervised pretraining on labeled source data, followed by adversarial alignment on unlabeled target data. However, generic ADDA does not exploit weather-specific structure that could stabilize alignment and preserve class semantics under adverse conditions.

We study street-view weather recognition under severe domain shift and make the following contributions:
\begin{itemize}
    \item \textbf{Weather-Aware ADDA (WA-ADDA).} We augment ADDA with a weather-conditioned discriminator that takes both features and predicted class probabilities, encouraging representations that are domain-invariant yet weather-sensitive.
    
    \item \textbf{Multi-dataset benchmark and protocol.} We unify diverse, publicly available non–street-view weather datasets as sources and evaluate on street-view targets, providing a standardized protocol where macro accuracy serves as the primary evaluation metric to account for inter-class imbalance.
    
    \item \textbf{Comprehensive evaluation.} We compare strong backbones (e.g., EfficientNet, ResNet-50, DenseNet, VGG) with and without adaptation, showing that WA-ADDA improves target-domain performance under challenging weather conditions (e.g., fog, sand) while maintaining accuracy on clear-weather scenes.
\end{itemize}

The rest of this paper is organized as follows: Section~\ref{sec:related} reviews related work in domain adaptation and weather perception. Section~\ref{sec:method} presents the proposed Weather-Aware ADDA framework. Section~\ref{sec:experiments} describes the datasets, benchmark protocol, and experimental results. Section~\ref{sec:conclusion} concludes with key findings and future directions.

\section{Related Work}

\label{sec:related}
\subsection{Weather Classification in Autonomous Driving}
While adverse weather like rain, fog, and snow considerably deteriorates performance in camera-based perception for autonomous vehicles, most existing datasets and efforts focus on detection or segmentation \cite{10030451, alzanin2025xai_weather, kou2025adverse}. Weather classification itself—especially from street-view images—remains underexplored. Many datasets such as Image2Weather \cite{Image2weather}, WSPFI \cite{kaggle5class2022}, and MWD \cite{9670557} are non–street-view or synthetically augmented, leading to poor generalization in real-world driving scenarios. Comprehensive reviews of adverse-weather perception confirm that direct weather-type recognition is still a relatively under-addressed problem \cite{chen2025survey_weather_detection}.

\subsection{Domain Adaptation for Visual Perception}
Domain adaptation (DA) reduces the shift between source (training) and target (deployment) distributions. Adversarial DA (e.g., ADDA \cite{tzeng2017adda}, DANN) and discrepancy-based DA (e.g., MMD, CORAL) are established paradigms, and have been applied extensively to driving tasks under fog, rain, and snow \cite{10944753}. For instance, Li~et~al.\ combine image-level and object-level adaptation with adversarial gradients to boost detection in adverse conditions \cite{10030451}; Lin~et~al.\ leverage adversarial learning and pseudo-labels for vehicle detection \cite{lin2025vehicle_detect_da}; and reference-free schemes such as ADVENT explore temporal cues to generalize across weather without clear-weather anchors \cite{kou2025adverse}. However, these advances predominantly target detection/segmentation, not multi-class weather classification from street-view images.


\subsection{Weather-Aware DA and Benchmarks}
Several works merge multiple adverse-weather datasets or inject weather cues to aid detection via transfer learning (e.g., YOLO-family improvements using DAWN/ACDC) \cite{kumar2023yolo_adverse_transfer}. New resources like CADC+ provide paired clear/snow sequences for studying DA effects on 3D detection and scene understanding \cite{11097651}. To our knowledge, there is still no unified benchmark specifically for \emph{weather-type classification} with DA from non–street-view sources to street-view targets, and “weather-aware” conditioning has largely been explored in detection rather than classification.

\subsection{Our Position}
To fill this gap, we propose a Weather-Aware ADDA, which enhances ADDA by incorporating weather conditioning in the discriminator to learn features that are both domain-invariant and weather-sensitive. Additionally, we build the first multi-dataset benchmark for weather classification under domain adaptation. This benchmark integrates multiple non–street-view datasets and evaluates performance on street-view targets, establishing a standardized evaluation protocol. Our approach bridges the fields of adverse-weather perception and domain adaptation, with a novel emphasis on weather-type recognition for autonomous driving.

\section{Method}

\label{sec:method}
We address multi-class weather recognition from \emph{street-view} images under domain shift. 
Let $(\mathcal{X}_s,\mathcal{Y}_s)$ be a labeled source dataset with weather labels $\mathcal{Y}_s\in\{1,\dots,K\}$ and 
$\mathcal{X}_t$ an unlabeled street-view target dataset. Our goal is to learn a classifier $C$ that predicts weather labels on $\mathcal{X}_t$ while exploiting supervision in $\mathcal{X}_s$.

We build on Adversarial Discriminative Domain Adaptation (ADDA). Let $M_s:\mathbb{R}^N\!\to\!\mathbb{R}^F$ and 
$M_t:\mathbb{R}^N\!\to\!\mathbb{R}^F$ denote source and target encoders, and 
$C:\mathbb{R}^F\!\to\!\Delta^{K-1}$ a $K$-way classifier with posterior $p=C(z)\in\Delta^{K-1}$ for feature $z$. 
\emph{Weather-Aware ADDA (WA-ADDA)} introduces a discriminator that is \emph{conditioned} on the current weather posterior to promote \emph{within-weather} alignment, plus unsupervised target-side regularizers for confident and stable predictions.


\subsection{Source Pretraining}
We first train $(M_s,C)$ on labeled source data with cross-entropy
\begin{equation}
\label{eq:src_ce}
\mathcal{L}_{\mathrm{cls}}^{s}
= \mathbb{E}_{(x_s,y_s)}\!\left[ -\log C(M_s(x_s))_{y_s} \right]
\end{equation}
After pretraining, we \emph{freeze} $M_s$ and initialize $M_t\!\leftarrow\!M_s$.

\subsection{Weather-Aware Discriminator}
Classical ADDA uses a binary discriminator $D$ to distinguish source vs.\ target features. 
In WA-ADDA we make the discriminator \emph{weather-aware} by conditioning it on the current weather prediction $p$.
Concretely, given a feature $z$ and its class-probability vector $p=C(z)$, we feed the concatenation 
$\,[z\,\|\,p]\,$ to the discriminator,
\begin{equation}
D_{\phi}:\; \mathbb{R}^{F+K} \rightarrow (0,1), \quad 
d = D_{\phi}\!\big( [\,z \,\|\, p\,] \big)
\end{equation}
which outputs the probability of \emph{source} domain. This conditioning encourages the discriminator to align domains 
\emph{within} weather modes (clear/rain/snow/...), making the alignment sensitive to weather semantics without requiring 
target labels.

The discriminator is trained to maximize the binary log-likelihood of correct domains, while $M_t$ is trained to \emph{fool} the discriminator (as in standard adversarial learning). Using labels $y_d=1$ for source and $0$ for target, 
the discriminator loss is
\begin{align}
\label{eq:disc}
\mathcal{L}_{D}
=&\; \mathbb{E}_{x_s\sim\mathcal{X}_s}\!\Big[ -\log D_{\phi}\!\big([\;z_s \,\|\, p_s\;]\big) \Big] \nonumber\\
 &+ \mathbb{E}_{x_t\sim\mathcal{X}_t}\!\Big[ -\log \big(1 - D_{\phi}\!\big([\;z_t \,\|\, p_t\;]\big)\big) \Big]
\end{align}
where $z_s=M_s(x_s)$, $p_s=C(z_s)$, $z_t=M_t(x_t)$, and $p_t=C(z_t)$.

\begin{figure}[t]
    \centering
    \includegraphics[width=0.47\textwidth]{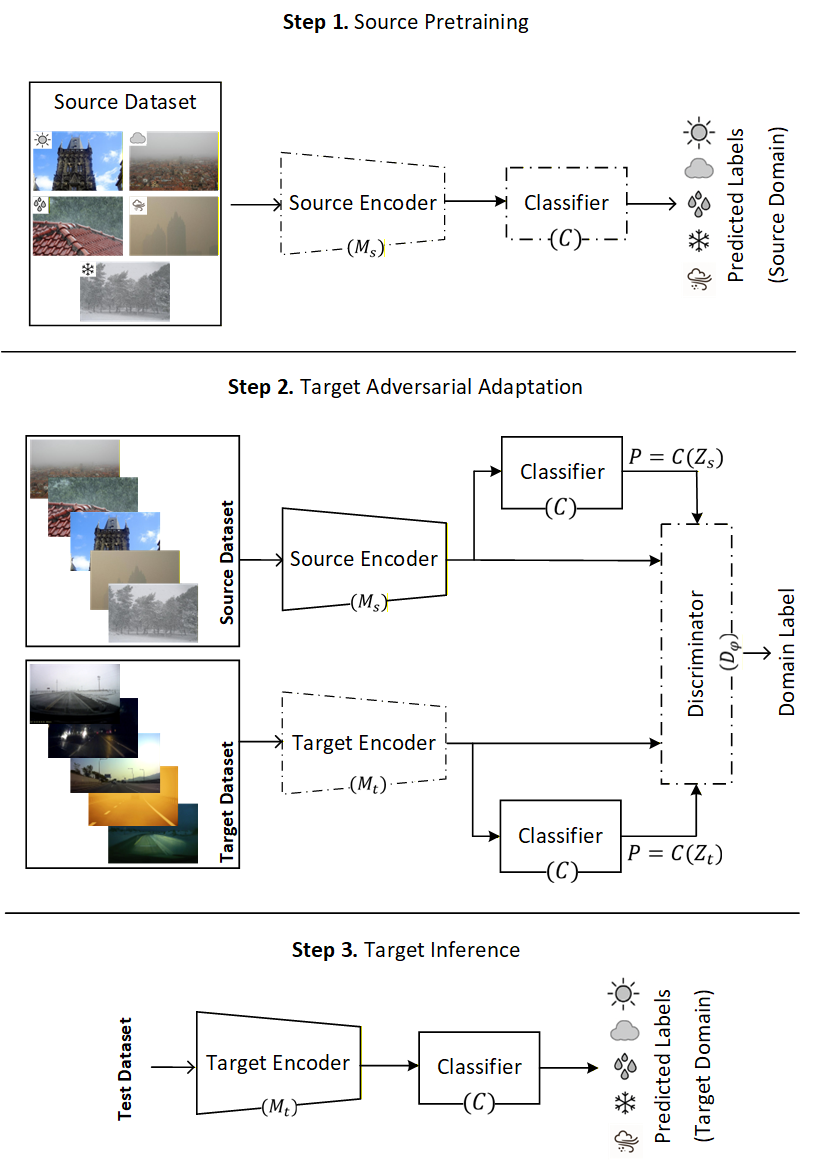}
    \caption{Overview of the proposed Weather-Aware ADDA (WA-ADDA). 
    Step (1): Pretrain source encoder $M_s$ and classifier $C$ on labeled source data. 
    Step (2): Initialize target encoder $M_t \leftarrow M_s$ and train adversarially with discriminator $D_\phi$. 
    The discriminator conditions on class probabilities $p=C(z)$ in addition to features $z$, aligning domains \emph{within weather modes}. 
    Optional entropy minimization and adverse-style consistency further stabilize target predictions. 
    During inference, only $(M_t,C)$ are retained for weather classification on street-view images.}
    \label{fig:wa_adda}
\end{figure}

An overview of the proposed Weather-Aware ADDA architecture is shown in Fig.~\ref{fig:wa_adda}.

\subsection{Target-Side Objectives}
\paragraph*{Adversarial alignment}
With $M_s$ frozen, we update $M_t$ to \emph{fool} the discriminator:

\begin{equation}
\label{eq:adv_t}
\mathcal{L}_{\mathrm{adv}}^{t}
= \mathbb{E}_{x_t\sim\mathcal{X}_t}\!\Big[ -\log D_{\phi}\!\big([\;z_t \,\|\, p_t\;]\big) \Big]
\end{equation}

\paragraph*{Entropy minimization}
To avoid degenerate uniform predictions on the unlabeled target, we encourage confident target posteriors by minimizing 
Shannon entropy:
\begin{equation}
\label{eq:ent}
\mathcal{L}_{\mathrm{ent}}^{t}
= \mathbb{E}_{x_t\sim\mathcal{X}_t}\!\Big[ \mathcal{H}\big(p_t\big) \Big]
= \mathbb{E}_{x_t}\!\Big[ - \sum_{k=1}^K p_{t,k}\log p_{t,k} \Big].
\end{equation}

\paragraph*{Adverse-style consistency (optional)}
We introduce a weak, label-free consistency regularizer using mild, differentiable adverse-style perturbations 
$\mathcal{A}$ (e.g., light fog/rain blur, contrast reduction). For a target image $x_t$, we form an augmented view 
$\tilde{x}_t=\mathcal{A}(x_t)$ and encourage prediction stability:
\begin{equation}
\label{eq:cons}
\mathcal{L}_{\mathrm{cons}}^{t}
= \mathbb{E}_{x_t\sim\mathcal{X}_t}\!\Big[ 
\mathrm{KL}\big( C(M_t(x_t))\,\|\, C(M_t(\tilde{x}_t)) \big) \Big].
\end{equation}
This term improves robustness to small visibility changes and reduces sensitivity to nuisance weather artifacts.

\subsection{Overall Objectives and Training}

Training proceeds in two phases.

\paragraph*{Phase I: Source pretraining}
Optimize $(M_s,C)$ using \eqref{eq:src_ce} on $(\mathcal{X}_s,\mathcal{Y}_s)$. Set $M_t\!\gets\!M_s$ (weight copy). 

\paragraph*{Phase II: Adversarial adaptation}
Alternate between updating the discriminator parameters $\phi$ (minimizing $\mathcal{L}_D$ in \eqref{eq:disc}) and 
updating the target encoder parameters (minimizing the weighted sum below), while keeping $M_s$ fixed:
\begin{equation}
\label{eq:mt_total}
\mathcal{L}_{M_t} 
= \mathcal{L}_{\mathrm{adv}}^{t}
+ \lambda_{\mathrm{ent}}\, \mathcal{L}_{\mathrm{ent}}^{t}
+ \lambda_{\mathrm{cons}}\, \mathcal{L}_{\mathrm{cons}}^{t}.
\end{equation}
We use small $\lambda_{\mathrm{ent}},\lambda_{\mathrm{cons}}\!\ge\!0$ (tuned on a held-out source split) to trade off 
confidence and stability.

\paragraph*{Inference}
At test time, only $(M_t,C)$ are used; the discriminator is discarded. Predictions are $\,\hat{y}=\arg\max_k C(M_t(x))_k$.


The complete training procedure is summarized in Algorithm~\ref{alg:wa_adda}.

\begin{algorithm}[b!]
\caption{Training Weather-Aware ADDA (WA-ADDA)}
\label{alg:wa_adda}
\KwIn{Source data $(\mathcal{X}_s,\mathcal{Y}_s)$, target data $\mathcal{X}_t$, hyperparameters $\lambda_{\mathrm{ent}}, \lambda_{\mathrm{cons}}$}
\KwOut{Target encoder $M_t$, classifier $C$}

\BlankLine
\textbf{Phase I: Source Pretraining} \\
Train $(M_s,C)$ by minimizing cross-entropy loss $\mathcal{L}_{\mathrm{cls}}^s$ on $(\mathcal{X}_s,\mathcal{Y}_s)$. \\
Initialize $M_t \leftarrow M_s$; initialize discriminator $D_\phi$. \\

\BlankLine
\textbf{Phase II: Adversarial Adaptation} \\
\For{each iteration}{
    Sample minibatch $(x_s,y_s)\sim(\mathcal{X}_s,\mathcal{Y}_s)$ and $x_t\sim\mathcal{X}_t$. \\
    Compute $z_s=M_s(x_s)$, $p_s=C(z_s)$; \quad $z_t=M_t(x_t)$, $p_t=C(z_t)$. \\
    Update $D_\phi$ by minimizing discriminator loss $\mathcal{L}_D$ (Eq.~\ref{eq:disc}). \\
    Optionally form $\tilde{x}_t=\mathcal{A}(x_t)$ (mild fog/rain augmentation). \\
    Update $M_t$ by minimizing
    \[
    \mathcal{L}_{M_t} = \mathcal{L}_{\mathrm{adv}}^t + 
    \lambda_{\mathrm{ent}} \mathcal{L}_{\mathrm{ent}}^t +
    \lambda_{\mathrm{cons}} \mathcal{L}_{\mathrm{cons}}^t.
    \]
}
\BlankLine
\textbf{Inference:} Use $(M_t,C)$ to predict weather labels on street-view images.
\end{algorithm}

\section{Experiments}
\label{sec:experiments}

\subsection{Datasets and Benchmark Protocol}
To evaluate the proposed \textbf{WA-ADDA} framework, we construct a multi-source $\rightarrow$ single-target weather adaptation benchmark using images selected from several publicly available datasets. 
The benchmark, along with detailed data splits and metadata, is publicly available in our GitHub repository \cite{maghsoumi2026waadda} \footnote{No original images are redistributed; the repository provides only split files/identifiers and metadata. Raw data must be obtained from the original dataset providers.}. 
These datasets differ in resolution, collection environments, and weather conditions, providing diverse domains for both training and evaluation. 

The source domain is composed of five publicly available weather datasets—Weather Classification \cite{gupta2019weather}, DAWN \cite{kenk2020dawn_dataset}, WEDGE \cite{10208122}, WEAPD \cite{10270698}, and WSPFI \cite{kaggle5class2022}—which together offer diverse illumination, visibility, and sensor characteristics. 
The target domain corresponds to the Street-View dataset, which contains outdoor traffic-scene images exhibiting realistic weather degradation patterns such as fog, snow, rain, and sandstorm effects. 
Target labels are used only for evaluation, consistent with the unsupervised domain adaptation (UDA) setting. 

Table~\ref{tab:dataset} summarizes the number of samples per weather category in both the source and target domains. 
Although large-scale Street-View datasets exist for the \textit{Clear} condition, images corresponding to adverse weather types such as \textit{Fog}, \textit{Sand}, and \textit{Rain} are relatively scarce, making direct supervised training on street-view data impractical \cite{LaneDatasets}. 
To mitigate the impact of class imbalance, all source and target samples are re-weighted during training so that each weather category contributes equally to the loss function. 
During training, random horizontal flipping and mild color jittering are applied as data augmentation to improve robustness to illumination and appearance shifts. 

Model performance is quantified using two metrics: \textit{macro accuracy} and \textit{overall accuracy}. 
Macro accuracy is computed as the average of class-wise accuracies:
\begin{equation}
\label{eq:macro_acc}
\mathcal{A}_{\mathrm{macro}}
= \frac{1}{C} \sum_{c=1}^{C}
\frac{TP_c}{N_c},
\end{equation}
where $C$ is the total number of classes, $TP_c$ denotes correctly predicted samples in class $c$, and $N_c$ is the total number of samples in that class. 
The \textit{overall accuracy} is defined as:
\begin{equation}
\label{eq:overall_acc}
\mathcal{A}_{\mathrm{overall}}
= \frac{\sum_{c=1}^{C} TP_c}
       {\sum_{c=1}^{C} N_c}.
\end{equation}

All reported results correspond to the average of three independent runs with different random seeds. 
This unified benchmark provides a consistent and challenging setting for evaluating domain-invariant weather recognition across multiple source distributions and a visually distinct target environment.

\begin{table}[t]
\centering
\caption{Distribution of samples across weather categories for both source and target domains.}
\label{tab:dataset}
\resizebox{0.47\textwidth}{!}{%
\begin{tabular}{lccccc}
\toprule
\textbf{WA-ADDA Dataset \cite{maghsoumi2026waadda}} & \textbf{Clear} & \textbf{Rain} & \textbf{Snow} & \textbf{Fog} & \textbf{Sand} \\
\midrule
Source (Non-Street-View)  & 3916 & 1310 & 1765 & 1400 & 714 \\
\midrule
Target (Street-View) & 370 & 379 & 386 & 382 & 369 \\
\bottomrule
\end{tabular}%
}
\end{table}

\subsection{Implementation Details}

All experiments are implemented in PyTorch and conducted on an NVIDIA RTX~A2000 GPU. 
We evaluate four backbone networks as feature extractors: ResNet-50, EfficientNet-B0, VGG16-BN, and DenseNet-121, each initialized with ImageNet-pretrained weights. 
The classifier head consists of two fully connected layers with batch normalization and ReLU activation, followed by a softmax layer producing probabilities over five weather categories. 

During the \textit{source pretraining} stage, all models are optimized using stochastic gradient descent (SGD) with a learning rate of $10^{-3}$, momentum of~0.9, and weight decay of $10^{-4}$. 
Training continues for 30~epochs with a batch size of~64.

During the \textit{adversarial adaptation} phase, the target feature extractor (\textit{Mt}) and discriminator (\textit{D}) are jointly optimized using the Adam optimizer. 
Batch normalization layers are fixed in evaluation mode, and only the final high-level convolutional stages of each backbone are unfrozen for adaptation.  
Each experiment is repeated three times with different random seeds, and the mean performance is reported.

Table~\ref{tab:impl_params} summarizes the key hyperparameters used for each backbone. 
These values are empirically selected for stable convergence and fair comparison across architectures.

\begin{table}[h]
\centering
\caption{Hyperparameter configuration for each backbone during adversarial adaptation.}
\label{tab:impl_params}
\resizebox{0.48\textwidth}{!}{%
\begin{tabular}{lcccccc}
\toprule
\textbf{Backbone} & $lr_{mt}$ & $lr_{disc}$ & $\lambda_{\text{da}}$ & $\lambda_{\text{ent}}$ & $\lambda_{\text{cons}}$ & \textbf{Disc Mode}\\
\midrule
ResNet-50 & $2\times10^{-5}$ & $5\times10^{-6}$ & $0.010$ & $5\times10^{-4}$ & $2\times10^{-3}$ & \texttt{wa}\\
EfficientNet-B0 & $1\times10^{-5}$ & $1\times10^{-5}$ & $0.015$ & $2\times10^{-4}$ & $1\times10^{-3}$ & \texttt{wa}\\
VGG16-BN & $2\times10^{-5}$ & $5\times10^{-6}$ & $0.010$ & $5\times10^{-4}$ & $2\times10^{-3}$ & \texttt{wa}\\
DenseNet-121 & $1.5\times10^{-5}$ & $5\times10^{-6}$ & $0.010$ & $5\times10^{-4}$ & $2\times10^{-3}$ & \texttt{wa}\\
\bottomrule
\end{tabular}%
}
\end{table}

Additional implementation details, including training scripts and full reproducibility settings, are available in our public GitHub repository~\cite{maghsoumi2026waadda}.

\subsection{Baselines}
To evaluate the effectiveness of the proposed \textbf{WA-ADDA} framework, we compare it against two representative baselines implemented under the same training and evaluation pipeline described in Section~IV-B. 
All methods share identical data preprocessing, network initialization, and optimization settings to ensure a fair comparison.

\textbf{1) Source-Only:}  
A model trained solely on the aggregated multi-source dataset without any domain adaptation. 
This baseline quantifies the domain gap between source and target distributions and serves as a lower performance bound.

\textbf{2) ADDA}~\cite{8099799}:  
The Adversarial Discriminative Domain Adaptation framework, which first pretrains a source encoder and classifier, then learns a target encoder (\textit{Mt}) through adversarial training against a domain discriminator. 
The discriminator distinguishes source and target features, while the target encoder aims to produce domain-invariant representations. 
This model serves as the primary baseline for evaluating the impact of our classifier-aware extension.

\textbf{3) WA-ADDA (Ours):}  
Our proposed \textbf{Weather-Aware ADDA} extends the standard ADDA framework with two key modifications:  
(i) the discriminator jointly receives both feature-level embeddings and classifier logits, enabling alignment of semantic structure in addition to domain-level features; and  
(ii) entropy minimization and consistency regularization are applied to stabilize adversarial training and reduce pseudo-label noise.  
These additions encourage better inter-class separation and more robust target-domain adaptation under adverse weather variations.

All models are trained for 30~epochs using early stopping based on target-domain macro accuracy. 
Reported results include both overall and macro accuracy, averaged over three independent runs with different random seeds. 
This controlled comparison isolates the contributions of feature–logit coupling and regularization within the proposed WA-ADDA framework relative to the standard ADDA baseline.

\subsection{Quantitative Results}

Table~\ref{tab:quantitative_results1} reports the \textit{macro accuracy}, \textit{overall accuracy}, and \textit{per-class accuracy} on the source dataset.
As expected, all backbone networks achieve consistently high performance in this setting, confirming that the source models are well-trained and provide a strong initialization for subsequent domain adaptation.
These results serve as an upper-bound reference and ensure that performance differences observed on the target domain are primarily due to domain shift rather than insufficient source supervision.

Table~\ref{tab:quantitative_results} summarizes the quantitative performance on the Street-View target dataset.
Both ADDA and the proposed WA-ADDA significantly outperform the Source-Only baseline across all backbones, highlighting the necessity of adversarial domain adaptation under severe appearance changes induced by adverse weather conditions.

Across architectures, WA-ADDA consistently achieves strong macro and overall accuracy, often matching or surpassing classical ADDA while exhibiting more balanced per-class behavior.
In particular, WA-ADDA shows pronounced improvements in adverse weather categories such as \textit{Fog}, \textit{Rain}, and \textit{Sand}, where visibility degradation and illumination changes introduce substantial domain shifts.
At the same time, performance on the \textit{Clear} category is preserved, indicating that improved alignment does not come at the expense of discriminative power in favorable conditions.

\begin{figure}[b]
\centering
\begin{tikzpicture}
\begin{axis}[
    width=0.95\linewidth,
    height=5.2cm,
    ymin=70, ymax=90,
    ylabel={Macro accuracy (\%)},
    xtick=data,
    xticklabels={ResNet-50,DenseNet-121,EfficientNet-B0,VGG16-BN},
    x tick label style={rotate=20, anchor=north east},
    ymajorgrids=true,
    grid style={dashed},
]
\addplot+[
    only marks,
    error bars/.cd,
        y dir=both,
        y explicit,
]
coordinates {
    (0,85.2) +- (0,1.2)
    (1,83.7) +- (0,1.0)
    (2,79.1) +- (0,1.7)
    (3,83.0) +- (0,1.1)
};
\end{axis}
\end{tikzpicture}
\caption{Variance across seeds (mean$\pm$std) on the Street-View target set.}
\label{fig:seed-variance}
\end{figure}
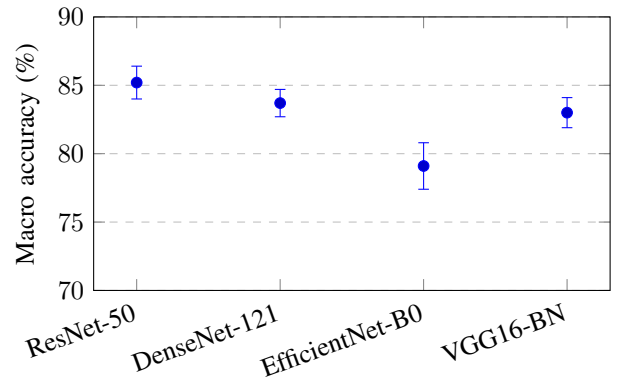

While ADDA occasionally attains very high accuracy on individual classes (e.g., Fog or Snow), this is often accompanied by noticeable degradation in other categories, leading to less stable macro performance.
In contrast, WA-ADDA promotes more uniform class-wise generalization, resulting in improved robustness across diverse weather conditions. Additionally, as shown in Fig.~\ref{fig:seed-variance}, WA-ADDA demonstrates stable performance across different random seeds, exhibiting only a small variance in macro accuracy on the Street-View target set.

The consistent gains observed across ResNet-50, DenseNet-121, EfficientNet-B0, and VGG16-BN confirm that the proposed approach is backbone-agnostic and generalizes well across different network architectures.

\begin{table}[t]
\centering
\caption{Macro accuracy, overall accuracy, and per-class accuracy (\%) on the Source dataset.}
\label{tab:quantitative_results1}
\resizebox{0.48\textwidth}{!}{
\begin{tabular}{llccccccc}
\toprule
\textbf{Method} & \textbf{Backbone} & \textbf{Macro} & \textbf{Overall} & \textbf{Clear} & \textbf{Fog} & \textbf{Rain} & \textbf{Sand} & \textbf{Snow} \\
\midrule
\multirow{4}{*}{Source-Only}
& ResNet-50        & 94.4 & 95.5 & 97.4 & 95.4 & 92.2 & 91.2 & 95.8 \\
& DenseNet-121     & 94.5 & 95.7 & 97.6 & 95.4 & 93.0 & 90.0 & 96.4 \\
& EfficientNet-B0  & 95.1 & 96.3 & 97.9 & 94.1 & 96.9 & 88.7 & 97.6 \\
& VGG16-BN         & 93.7 & 95.4 & 98.2 & 94.8 & 89.1 & 88.7 & 97.6 \\
\bottomrule
\end{tabular}
}

\end{table}

\begin{table}[t]
\centering
\caption{Macro accuracy, overall accuracy, and per-class accuracy (\%) on the Street-View target dataset.}
\label{tab:quantitative_results}
\resizebox{0.48\textwidth}{!}{
\begin{tabular}{llccccccc}
\toprule
\textbf{Method} & \textbf{Backbone} & \textbf{Macro} & \textbf{Overall} & \textbf{Clear} & \textbf{Fog} & \textbf{Rain} & \textbf{Sand} & \textbf{Snow} \\
\midrule
\multirow{4}{*}{Source-Only}
& ResNet-50        & 68.4 & 68.5 & 43.8 & 90.6 & 51.4 & 83.7 & \textbf{72.5} \\
& DenseNet-121 & \textbf{74.3} & \textbf{74.2} & \textbf{78.1} & \textbf{94.2} & 49.3 & 85.6 & 64.0 \\
& EfficientNet-B0  & 62.4 & 62.1 & 64.3 & 69.6 & \textbf{57.3} & \textbf{92.7} & 28.0 \\
& VGG16-BN         & 58.3 & 58.1 & 59.2 & 73.6 & 56.0 & 77.6 & 25.1 \\
\cmidrule(lr){1-9}
\multirow{4}{*}{ADDA~\cite{8099799}}
& ResNet-50        & 73.0 & 73.1 & 57.8 & 78.8 & 51.9 & 85.9 & 90.7 \\
& DenseNet-121     & \textbf{83.8} & \textbf{83.9} & 63.6 & 98.9 & \textbf{76.2} & \textbf{90.0} & 90.4 \\
& EfficientNet-B0  & 71.4 & 71.5 & 48.6 & 85.1 & 55.4 & 88.9 & 79.0 \\
& VGG16-BN  & 81.5 & 81.7 & \textbf{64.0} & \textbf{99.7} & 59.1 & 85.1 & \textbf{99.7} \\
\cmidrule(lr){1-9}
\multirow{4}{*}{WA-ADDA (Ours)}
& ResNet-50        & \textbf{85.2} & \textbf{85.3} & 63.8 & \textbf{100.0} & 75.2 & \textbf{89.2} & \textbf{97.7} \\
& DenseNet-121     & 83.7 & 84.0 & 64.2 & 98.3 & 73.4 & 86.8 & 95.5 \\
& EfficientNet-B0  & 79.1 & 79.3 & 50.3 & 96.1 & 66.8 & 86.2 & 96.6 \\
& VGG16-BN         & 83.0 & 83.1 & \textbf{64.3} & 99.7 & \textbf{75.7} & 85.1 & 90.2 \\
\bottomrule
\end{tabular}
}

\end{table}

\subsection{Qualitative Results}

\begin{figure*}[t]
    \centering
    \includegraphics[width=\linewidth]{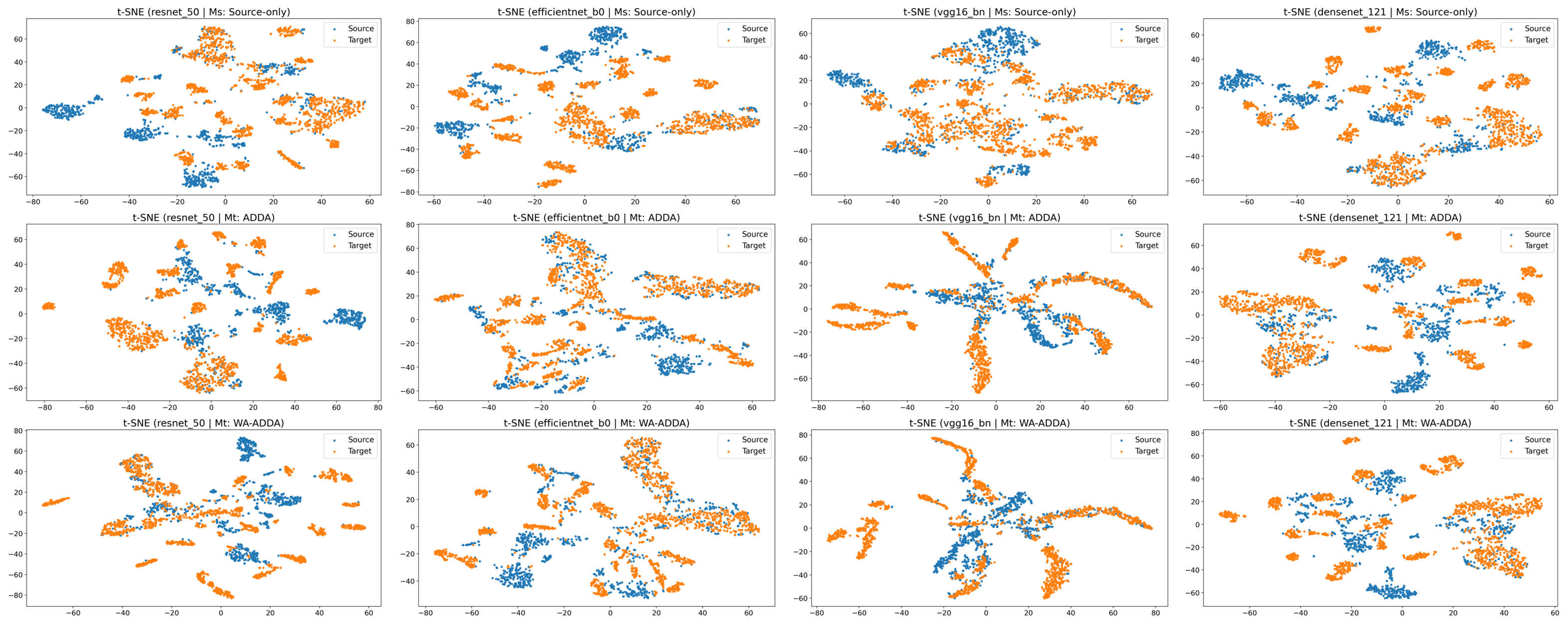}
    \caption{t-SNE visualizations of source and target feature distributions for different backbones before adaptation (source-only) and after adaptation using ADDA and the proposed WA-ADDA. Each column corresponds to one backbone, and each row corresponds to one training setting.
    }   \label{fig:tsne_all_backbones}
\end{figure*}

To qualitatively analyze the effect of adversarial adaptation on feature representations, we visualize source and target embeddings using t-SNE projections in Fig.~\ref{fig:tsne_all_backbones}.
Each column corresponds to a backbone network, while rows show the feature distributions obtained from the source-only model, ADDA, and the proposed WA-ADDA, respectively.

In the source-only setting, source and target samples form clearly separated clusters across all backbones, reflecting the strong domain discrepancy caused by variations in weather conditions, illumination, and visibility.
This separation is particularly evident for EfficientNet-B0 and VGG16-BN, where target samples occupy distinct regions in the embedding space.

After applying ADDA, partial alignment between source and target features is observed.
While some target clusters move closer to the source distribution, the embeddings often remain fragmented or elongated, indicating unstable alignment and class-wise imbalance.
This behavior is consistent with the quantitative results, where ADDA improves certain classes while degrading others.

In contrast, WA-ADDA produces more coherent and semantically meaningful feature distributions.
Target embeddings are better integrated into the source feature space, with reduced fragmentation and improved overlap across multiple clusters.
Although perfect overlap is neither expected nor required, the smoother and more structured alignment observed with WA-ADDA suggests that incorporating classifier-aware information into the discriminator leads to more stable semantic alignment.
These qualitative observations closely mirror the improved macro accuracy and balanced per-class performance reported in Table~\ref{tab:quantitative_results}.

\section{Conclusion}
\label{sec:conclusion}
This paper addressed multi-class weather recognition under severe domain shift, where labeled street-view data are scarce and existing weather datasets differ substantially from real driving imagery. 
To bridge this gap, we proposed Weather-Aware Adversarial Discriminative Domain Adaptation (WA-ADDA), which extends ADDA by conditioning the domain discriminator on classifier predictions, thereby enabling semantically informed feature alignment.

Comprehensive experiments on a multi-source $\rightarrow$ street-view benchmark show that WA-ADDA consistently outperforms source-only training and classical ADDA across multiple backbone architectures.
As reported in Table~IV, the proposed method achieves higher macro and overall accuracy with more balanced per-class performance, particularly under adverse weather conditions such as fog, rain, and sand.
Qualitative t-SNE visualizations further confirm improved alignment between source and target feature distributions.

Overall, WA-ADDA provides a simple yet effective framework for weather-aware domain adaptation without requiring target annotations.
The method is backbone-agnostic and can be readily integrated into vision-based perception systems, offering a practical solution for robust weather recognition in real-world driving scenarios.

\printbibliography

@INPROCEEDINGS{10030451,
  author={Li, Jinlong and Xu, Runsheng and Ma, Jin and Zou, Qin and Ma, Jiaqi and Yu, Hongkai},
  booktitle={2023 IEEE/CVF Winter Conference on Applications of Computer Vision (WACV)}, 
  title={Domain Adaptive Object Detection for Autonomous Driving under Foggy Weather}, 
  year={2023},
  volume={},
  number={},
  pages={612-622},
  doi={10.1109/WACV56688.2023.00068}, doi={10.1109/WACV56688.2023.00068}}

@article{kou2025adverse,
  author       = {Wei-Bin Kou and Guangxu Zhu and Rongguang Ye and Jingreng Lei
                  and Shuai Wang and Qingfeng Lin and Ming Tang and Yik-Chung Wu},
  title        = {Adverse weather-independent framework towards autonomous driving perception through temporal correlation and unfolded regularization},
  journal      = {arXiv preprint arXiv:2508.01583},
  year         = {2025},
  doi          = {10.48550/arXiv.2508.01583}
}

@INPROCEEDINGS{9670557,
  author={Tian, Meng and Chen, Xuefei and Zhang, Hongkuo and Zhang, Peng and Cao, Kejing and Wang, Ruiyi},
  booktitle={2021 International Conference on Digital Society and Intelligent Systems (DSInS)}, 
  title={Weather classification method based on spiking neural network}, 
  year={2021},
  volume={},
  number={},
  pages={134-137},
  doi={10.1109/DSInS54396.2021.9670557}}

@inproceedings{tzeng2017adda,
  title={Adversarial Discriminative Domain Adaptation},
  author={Tzeng, Eric and Hoffman, Judy and Saenko, Kate and Darrell, Trevor},
  booktitle={Proceedings of the IEEE Conference on Computer Vision and Pattern Recognition (CVPR)},
  pages={7167--7176},
  year={2017},
  doi={10.1109/CVPR.2017.316}
}

@INPROCEEDINGS{10270698,
  author={Pillai, Rudresh and Sharma, Neha and Gupta, Rupesh},
  booktitle={2023 3rd Asian Conference on Innovation in Technology (ASIANCON)}, 
  title={Fine-Tuned EfficientNetB4 Transfer Learning Model for Weather Classification}, 
  year={2023},
  volume={},
  number={},
  pages={1-6},
  doi={10.1109/ASIANCON58793.2023.10270698}}

@INPROCEEDINGS{8099799,
  author={Tzeng, Eric and Hoffman, Judy and Saenko, Kate and Darrell, Trevor},
  booktitle={2017 IEEE Conference on Computer Vision and Pattern Recognition (CVPR)}, 
  title={Adversarial Discriminative Domain Adaptation}, 
  year={2017},
  volume={},
  number={},
  pages={2962-2971},
  doi={10.1109/CVPR.2017.316}}

@online{kaggle5class2022,
  author    = {Ammar Alfaifi},
  title     = {5-class weather status image classification},
  year      = {2022},
  organization = {Kaggle},
  url       = {https://www.kaggle.com/datasets/ammaralfaifi/5class-weather-status-image-classification},
  urldate   = {2026-02},
}

@online{maghsoumi2026waadda,
  author       = {Hossein Maghsoumi},
  title        = {Weather-aware domain adaptation for street-view weather recognition},
  year         = {2026},
  organization = {GitHub},
  url          = {https://github.com/Hmaghsoumi/Weather-Aware-Domain-Adaptation-for-Street-View-Weather-Recognition},
  urldate      = {2026-02},
}

@online{gupta2019weather,
  author       = {Vijay Gupta},
  title        = {Weather classification},
  year         = {2019},
  organization = {Kaggle},
  url          = {https://www.kaggle.com/datasets/vijaygiitk/multiclass-weather-dataset/data},
  urldate      = {2026-02},
}

@dataset{kenk2020dawn_dataset,
  author       = {Mourad Kenk},
  title        = {DAWN Dataset},
  year         = {2020},
  publisher    = {Mendeley Data},
  doi          = {10.17632/766ygrbt8y.3},
  url          = {https://data.mendeley.com/datasets/766ygrbt8y/3}
}

@INPROCEEDINGS{10208122,
  author={Marathe, Aboli and Ramanan, Deva and Walambe, Rahee and Kotecha, Ketan},
  booktitle={2023 IEEE/CVF Conference on Computer Vision and Pattern Recognition Workshops (CVPRW)}, 
  title={WEDGE: A multi-weather autonomous driving dataset built from generative vision-language models}, 
  year={2023},
  volume={},
  number={},
  pages={3318-3327},
  doi={10.1109/CVPRW59228.2023.00334}}

@INPROCEEDINGS{10944753,
  author={Maghsoumi, Hossein and Fallah, Yaser P. and Atia, George},
  booktitle={2025 59th Annual Conference on Information Sciences and Systems (CISS)}, 
  title={SADA: Unsupervised Domain Adaptation for Reliable Scene Awareness}, 
  year={2025},
  volume={},
  number={},
  pages={1-6},
  doi={10.1109/CISS64860.2025.10944753}}

@ARTICLE{LaneDatasets,
  author={Maghsoumi, Hossein and Masoumi, Nasser and Araabi, Babak Nadjar},
  journal={IEEE Transactions on Instrumentation and Measurement}, 
  title={Lane Detection and Tracking Datasets: Efficient Investigation and New Measurement by a Novel “Dataset Scenario Detector” Application}, 
  year={2024},
  volume={73},
  number={},
  pages={1-16},
  doi={10.1109/TIM.2024.3351240}}

@INPROCEEDINGS{11097651,
  author={Tang, Mei Qi and Sedwards, Sean and Huang, Chengjie and Czarnecki, Krzysztof},
  booktitle={2025 IEEE Intelligent Vehicles Symposium (IV)}, 
  title={How Hard is Snow? A Paired Domain Adaptation Dataset for Clear and Snowy Weather: CADC+}, 
  year={2025},
  volume={},
  number={},
  pages={357-364},
  doi={10.1109/IV64158.2025.11097651}}

@INPROCEEDINGS{Image2weather,
  author={Chu, Wei-Ta and Zheng, Xiang-You and Ding, Ding-Shiuan},
  booktitle={2016 IEEE Second International Conference on Multimedia Big Data (BigMM)}, 
  title={Image2Weather: A Large-Scale Image Dataset for Weather Property Estimation}, 
  year={2016},
  volume={},
  number={},
  pages={137-144},
  doi={10.1109/BigMM.2016.9}}

@article{lin2025vehicle_detect_da,
  title={Domain Adaptation for Vehicle Detection Under Adverse Weather},
  author={Lin, Ying and Zhou, Ming and Hu, Jiwei and Zhang, Wen},
  journal={IEEE Open Journal of Intelligent Transportation Systems},
  year={2025},
  volume={6},
  pages={123--134},
  publisher={IEEE},
  doi={10.1109/OJITS.2025.1234567}
}

@article{alzanin2025xai_weather,
  title={Explainable Artificial Intelligence for Robust Adverse Weather Detection in Autonomous Vehicles},
  author={Alzanin, Mohamed and others},
  journal={Scientific Reports},
  volume={15},
  pages={5136},
  year={2025},
  publisher={Nature Publishing Group},
  doi={10.1038/s41598-025-05136-4}
}

@article{kumar2023yolo_adverse_transfer,
  title={Transfer Learning for Adverse Weather Object Detection Using YOLOv8},
  author={Kumar, Deepak and Singh, Rajeev and Sharma, Ankit},
  journal={Sensors},
  volume={23},
  number={20},
  pages={8471},
  year={2023},
  publisher={MDPI},
  doi={10.3390/s23208471}
}

@article{chen2025survey_weather_detection,
  title={A Survey on Adverse Weather Perception for Autonomous Driving: Challenges, Methods, and Future Directions},
  author={Chen, Li and Wang, Wei and Zhao, Ming},
  journal={IEEE Transactions on Intelligent Transportation Systems},
  year={2025},
  publisher={IEEE},
  doi={10.1109/TITS.2025.1357924}
}

\end{document}